\documentclass{article}

\usepackage{arxiv}

\usepackage[utf8]{inputenc} 
\usepackage[T1]{fontenc}    
\usepackage{hyperref}       
\usepackage{url}            
\usepackage{booktabs}       
\usepackage{amsfonts}       
\usepackage{nicefrac}       
\usepackage{microtype}      
\usepackage{lipsum}
\usepackage{multirow}
\usepackage{makecell}
\usepackage{graphicx}
\usepackage{caption}
\usepackage[page]{appendix}

\graphicspath{ {include/} }

\title{Combining neural and knowledge-based approaches to Named Entity Recognition in Polish}

\author{
  Sławomir Dadas \\
  \emph{National Information Processing Institute, Warsaw, Poland}\\
  \texttt{slawomir.dadas@opi.org.pl} \\
}

\begin{document}
\maketitle

\begin{abstract}
Named entity recognition (NER) is one of the tasks in natural language processing that can greatly benefit from the use of external knowledge sources. We propose a named entity recognition framework composed of knowledge-based feature extractors and a deep learning model including contextual word embeddings, long short-term memory (LSTM) layers and conditional random fields (CRF) inference layer. We use an entity linking module to integrate our system with Wikipedia. The combination of effective neural architecture and external resources allows us to obtain state-of-the-art results on recognition of Polish proper names. We evaluate our model on data from PolEval 2018\footnote{\url{http://poleval.pl}} NER challenge on which it outperforms other methods, reducing the error rate by 22.4\% compared to the winning solution. Our work shows that combining neural NER model and entity linking model with a knowledge base is more effective in recognizing named entities than using NER model alone.
\end{abstract}


\section{Introduction}
Named entity recognition (NER) is a problem of finding and classifying instances of named entities in text. NER systems are usually designed to detect entities from a pre-defined set of classes such as person names, temporal expressions, organizations, addresses. Such methods can be used independently but they are often one of the first steps in a complex natural language understanding (NLU) workflows involving multiple models. Therefore the performance of NER systems can affect the performance of NLU downstream tasks. The problem of named entity recognition is challenging because often both contextual information and domain knowledge are required to accurately recognize and categorize named entities. 

\subsection{Prior work}
Popular approach to named entity recognition is to train a sequence labeling model, i.e. a machine learning model that assigns a label to each word in a sentence indicating whether that word is a part of named entity or not. In the past few years, methods based on neural networks were the dominant solutions to this problem. \citet{collobert2008unified} showed that it is possible to train effective sequence labelling models with neural architecture involving word embeddings. Later, other word embedding approaches became popular, notably Word2Vec \citep{mikolov2013distributed}, GloVe \citep{pennington2014glove} and FastText \citep{bojanowski2017enriching}. Deep learning NER systems utilised those models as well as LSTM layers (long short-term memory) for encoding the sentence level information and CRF (conditional random fields) inference layer \citep{DBLP:journals/corr/HuangXY15, dos2015boosting, lample2016neural, ma2016end}. Some studies highlighted the importance of character level features by integrating additional character based representations. To this end, CNNs (convolutional neural networks) \citep{dos2015boosting, ma2016end, chiu2016named} or bidirectional LSTMs \citep{lample2016neural, rei2017semi, liu2017empower, peters2017semi} were used. Most recently, state-of-the-art NER systems employed word representations based on pre-trained language models, either replacing classic word embedding approaches \citep{peters2017semi, peters2018deep} or using both representations jointly \citep{akbik2018coling}.

In Polish, early tools for named entity recognition were based on heuristic rules and pre-defined grammars \citep{piskorski2005named, gralinski2009named} or CRF models with hand-crafted features \citep{waszuk2012nerf, radziszewski2013tiered, marcinczuk2012optimizing, marcinczuk2013liner2}. \citet{pohl2013knowledge} used OpenCyc and Wikipedia to build purely knowledge-based NER system. Only recently methods involving deep learning were introduced \citep{borchmann2018lstmcrf, marcinczuk2018poldeepner}.

While modern named entity recogntion methods have made considerable progress in exploiting contextual information and long term dependencies in text, in some cases it is not sufficient to accurately recognize a named entity. When the context does not provide enough information, model should be able to use external knowledge to help with the detection and classification. Such a need exists, for example, in the case of abbreviations or highly ambiguous phrases that can refer to several different entities. Therefore, we believe that the problem of integrating knowledge sources with NER models should be explored. In this work, we focus on named entity recognition for Polish language. We show how such model can be integrated with Wikipedia and how can we improve its performance by using an external knowledge base. The method proposed in this publication may also be used in other languages.

\subsection{Contributions} 
Our contributions are the following: 1) We propose a named entity recognition system for Polish that combines deep learning architecture with knowledge-based feature extractors, achieving state-of-the art results for this task. 2) We propose a method utilizing an entity linking model based on Wikipedia to improve the accuracy of named entity recognition. Additionally, we release a tool for efficient labeling of Wikipedia's articles. 3) We make the source code of our method available, along with pre-trained models for NER, pre-trained Polish Word2Vec \citep{mikolov2013distributed} embeddings and ELMo \citep{peters2018deep} embeddings, labeled data set of articles from Polish Wikipedia and two lexicons.

\section{Problem description and system architecture}
\label{sec:architecture}
In this section, we describe the problem of Named Entity Recognition for the Polish language, following the guidelines of the National Corpus of Polish (NKJP) \citep{przepiorkowski2012narodowy}. We introduce the challenges that arise from this task. Then, we present the general architecture of our system and show how it addresses those challenges. Finally, we describe language resources and external modules used by the system.

\subsection{Problem description}
The National Corpus of Polish (NKJP) is one of the largest text corpora of Polish language. A part of the corpus, so called \emph{"one million subcorpus"}, has been manually annotated which allows to use it as a training set for many natural language processing tasks. Among others, the subcorpus contains annotations of named entities from several entity categories, which in turn may be further divided into subcategories. Therefore, our task is to detect and classify any mention of a named entity in text, assigning the correct category and subcategory if applicable. NKJP identifies the following set of categories:

\begin{itemize}
\item \textbf{persName} - This category includes names of real and fictional people. It is divided into three subcategories: \emph{forename}, \emph{surname} and \emph{addName}.
\item \textbf{orgName} - The category covers names of organizations, institutions, companies, government authorities, bands, sport teams, military groups etc. Additionally, a metonymic use of other words can also be classified as \emph{orgName} i.g. the use of word \emph{"Poland"} when referring to \emph{"Polish government"} (\emph{"Poland votes against controversial copyright rules."}).
\item \textbf{geogName} - This category contains historical and geographical regions, natural structures such as rivers, mountains, lakes, islands, astronomical objects including planets, stars, galaxies. It also contains man made structures such as buildings, streets, squares, parks and other urban objects. Occasionally, organization names can be classified as \emph{geogName} when referring to the name of the building (\emph{"Take me to the European Parliament."}).
\item \textbf{placeName} - Category covers geopolitical names and is further divided into five subcategories: \emph{bloc} subcategory includes groups of countries (e.g. \emph{"EU", "NATO"}), \emph{country} includes single countries and their synonyms, \emph{region} covers administrative divisions larger than a city but smaller than a country, \emph{settlement} includes cities, towns and villages, \emph{district} includes administrative divisions smaller than a settlement. Based on the context of a sentence, names that are usually classified as \emph{placeName} can be either an \emph{orgName} (\emph{"European Union voices serious concerns."}) or a \emph{geogName} (\emph{"For I will punish them that dwell in the land of Egypt, as I have punished Jerusalem."}).
\item \textbf{date} and \textbf{time} - Two categories covering date and time related temporal expressions.
\end{itemize}

Although the structure of the National Corpus of Polish in the context of named entity recognition requires a few task specific tweaks to address such issues as nested labels, overlapping entities, fragmented entities and derived entities, our model can still be easily adapted to other sequence labeling tasks, in particular to named entity recognition in other languages. We decided to resolve task specific issues with simple solutions that would not require to change the core model architecture. Here, we provide a brief explanations of those problems and our solutions.
\paragraph{Nested labels} NKJP defines main categories and subcategories for named entities. Out of six main categories, two contain subcategories. For \emph{placeName}, every instance of a named entity can be assigned exactly one of its subcategories. For \emph{persName}, each word in a named entity can possibly have different subcategory, or no subcategory at all. Our solution to this problem was to train two models: one for predicting main categories and another for predicting subcategories. The architecture of both models is identical, except the model for subcategories takes another one-hot feature vector as an input which is the output label of its parent model (the main category label). In this work, we describe the main model only since we use exactly the same training procedure and the same hyperparameters for the subcategory model.
\paragraph{Fragmented entities} In the NKJP data set, there are examples of discontinuous entities i.e. parts of the same named entity are separated by one or more tokens. Since the number of such cases was small, we have decided to ignore this issue by transforming fragmented entities either to a single continuous entity (when fragments were separated by single word, we included the word in the entity) or to separate entities (when the gap was longer than a single word).
\paragraph{Derived entities} There is a concept of \emph{derived entity} in the context of \emph{placeName} category in the NKJP dataset. Derived entity can be either a name of the inhabitant of the place e.g. country or city, or an adjective related this place. Our neural model does not address this issue at all - derived named entities are treated just like any other instance of \emph{placeName}. Instead, we added a post-processing step based on a lexicon of inhabitants and relational adjectives that is responsible for detecting those cases and tagging them with additional label.
\paragraph{Overlapping entities} Given an example named entity of \emph{"Johns Hopkins Hospital in Baltimore"}, it may be labeled as \emph{geogName} or \emph{orgName} depending on the sentence context. However, fragments of this entity can also be labeled as \emph{persName} (\emph{"Johns Hopkins"}) and \emph{placeName-settlement} (\emph{"Baltimore"}). There is a number of similar cases in the NKJP data set, where fragments of named entity are named entities themselves. That makes this task a multi-label classification problem where most samples are assigned a single label and small number of edge cases have two labels. To avoid transforming our model to multi-label classifier, we decided to move smaller overlapping entities to subcategory model i.e. we train the subcategory model as they were subcategories of the longer entity. It doesn't solve all cases - some cannot be properly labeled with this approach - but the model is able to learn most of them.

\subsection{System architecture}

\begin{figure}
  \centering
  \includegraphics[scale=0.6]{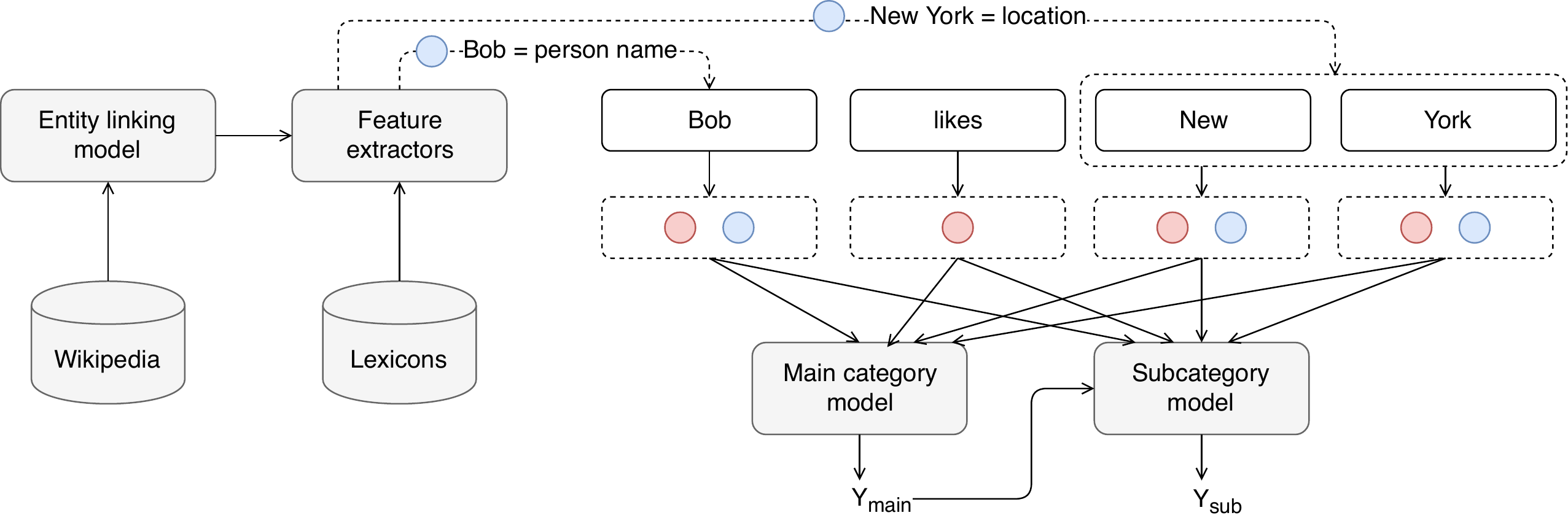}
  \caption{High level architecture of our named entity recognition system. Entity linking and feature extractor modules are responsible for enriching input with information from lexicons and Wikipedia. Input sequences are then sent to two neural models, predicting main categories and subcategories of named entities. A sequence of predicted main entity classes $Y_{main}$ is used as an input to subcategory model which outputs a sequence of entity subclasses $Y_{sub}$.}
  \label{fig:architecture_full}
\end{figure}

\begin{figure}
  \centering
  \includegraphics[scale=0.6]{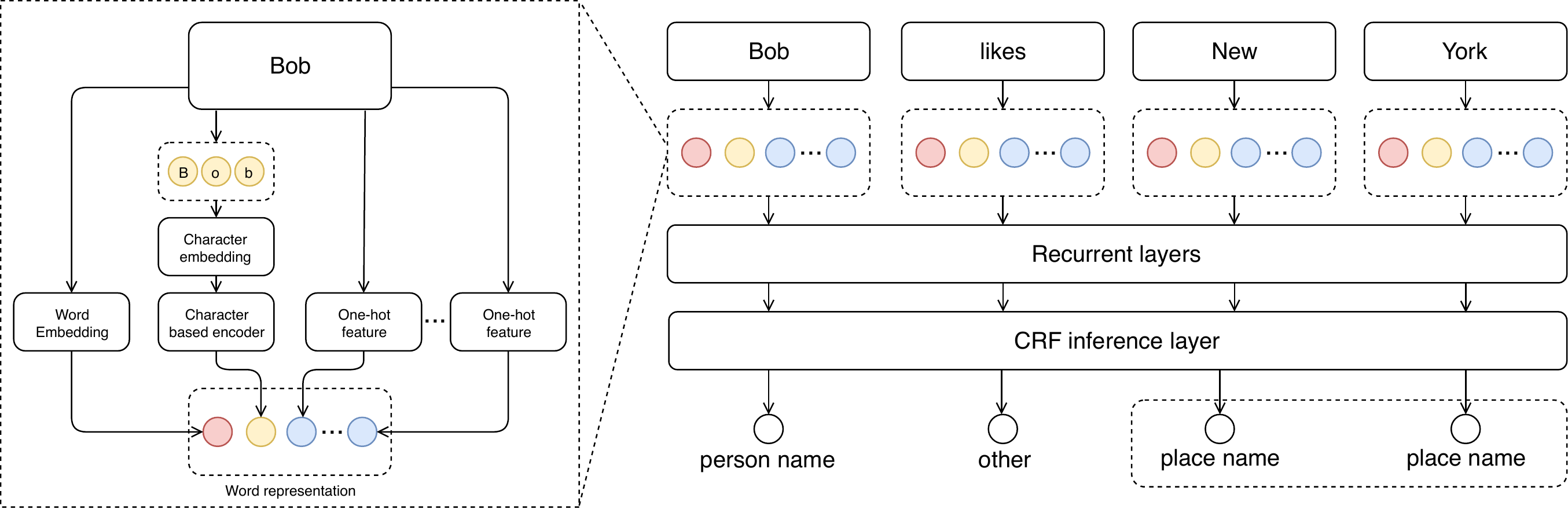}
  \caption{An architecture of our neural named entity recognition model. The model takes word sequence as an input, in this example \emph{"Bob likes New York"}. For each word in the sequence, a vectorized word representation is built. Figure on the left shows the structure of the module responsible for building word representations. Different components of the resulting vector are represented with different colors - red for word embedding, yellow for character encoder, blue for one-hot feature modules. On the right, complete architecture is shown, featuring recurrent layers and CRF inference layer.}
  \label{fig:architecture_model}
\end{figure}

Figure \ref{fig:architecture_full} shows a general architecture of our system. The system consists of four modules: an entity linking model based on Wikipedia (described in Section \ref{sec:wikipedia}), a feature extractors module (which integrates the aforementioned entity linker and a number of lexicons) and two deep learning models. Each word or word n-gram in an input sentence is first preprocessed by feature extractors that assign additional labels to it. The extractors used by our system are described in the next subsection. The enriched sequence of words is then used as an input to the neural models. An architecture of such model is shown in Figure \ref{fig:architecture_model}. Our model is similar to other recent deep learning NER architectures \citep{DBLP:journals/corr/HuangXY15, dos2015boosting, lample2016neural, ma2016end}. First, a vectorized representation of word is constructed by concatenating an output of pre-trained word embedding, a trainable character level encoder and a set of one-hot vectors from feature extraction module. Next, a hidden word representation is computed by a number of bidirectional LSTM layers. Finally, this representation is sent to a CRF output layer, which is responsible for predicting a sequence of labels $Y$ that maximizes the probability $P(Y|X)$, where $X$ is the sequence of word vectors. Two such models are utilised by our system, resulting in two output sequences: $Y_{main}$ for main categories and $Y_{sub}$ for subcategories of named entities. In order to correctly resolve multi word entities, we use BIO (beginning-inside-outside) tagging scheme as the output format of $Y_{main}$ and $Y_{sub}$ sequences.

\subsection{Feature extractors}
To support the detection of named entities, our system uses a number of additional feature extractors. Some of the features described below use static language resources, other are based on heuristic rules. The Wikipedia module deserves additional attention since it employs an external standalone service based on the Wikipedia-Miner project \citep{milne2013open}, using the Polish Wikipedia dump as a data source. Here, we present a short descriptions of all extractors, more information on the process of integrating our system with Wikipedia can be found in the Section \ref{sec:wikipedia}. Although the implementation of those feature extractors can vary in complexity, the purpose of each module is to assign a label to a word in the input sequence. Such label is then encoded as a one-hot vector and the vector is concatenated to produce the final word representation, consisting of the word embedding, character-based word representation and all the one-hot vectors provided by the feature modules. The full list of one-hot features used by our system includes:
\begin{itemize}
\item \textbf{Capitalization feature} - 
Following other works on Named Entity Recognition \citep{reimers2017optimal,DBLP:journals/corr/HuangXY15}, we add a capitalization feature providing information about the casing of the input. The feature can take one of eight labels: \emph{numeric} if all characters are digits, \emph{mostly\_numeric} if more than 50\% of characters are digits, \emph{upper} if all letters are upper case, \emph{lower} if all letters are lower case, \emph{title} if only the initial character is upper case, \emph{contains\_digit} if the word contains digit, \emph{mixed} when the word is a mix or upper and lower case characters, \emph{other} when none of the above is true.
\item \textbf{PoliMorf} -
PoliMorf \citep{wolinski2012polimorf} is an open-source Polish morphological dictionary containing over 7 million word forms assigned to over 200 word categories. In addition to the word categories, each entry in the dictionary contains a word lemma, part-of-speech tag and optional supplementary tags describing the word.
\item \textbf{PersonNames (lexicon of Polish, English and foreign names)} -
The lexicon contains about 346 thousand first and last names crawled from various Internet sources. Each name in the dictionary is labeled either as a Polish, English or foreign (other). Additional label indicates if a name is also a common word in Polish or English.
\item \textbf{GPNE (Gazetteer for Polish Named Entities)} -
The gazetteer has been created for the task of Named Entity Recognition in Polish and over the years it has been used by several NER systems i.e. the SProUT platform \citep{piskorski2004shallow, piskorski2005named} or NERF \citep{waszuk2012nerf}. It includes forenames and surnames, geographical names, institution names, country names and their derivatives and parts of temporal expressions among others.
\item \textbf{NELexicon2} -
This resource is a lexicon of named entities, created and maintained in Wroclaw University of Technology \citep{marcinczuk2013liner2}. Most of the entries have been extracted automatically from several sources such as Wikipedia infoboxes or Wiktionary. It contains more than 2.3 million names.
\item \textbf{Wikipedia (Entity Linking Module)} - While other features are based on static resources, this module is a contextual classifier which performs the task of entity linking i.e. recognizing and disambiuating mentions of entities from Wikipedia in text. After the linking phase, each found entity is assigned a label and the label is used as an input to the system. To make this possible, we have developed a tool for efficient labeling of Wikipedia articles. Using this tool, we managed to label 1.1 million articles in Polish Wikipedia (about 90\% of all articles). 
\item \textbf{Supplementary lexicon (Extras)} -
This lexicon has been prepared by us specifically for PolEval Named Entity Recognition task. Our intention was to create an additional resource containing classes of words that were not covered by other dictionaries but were essential for a successful NER system, in accordance with the guidelines of the National Corpus of Polish (NKJP). More specifically, the dictionary consists mostly of the names of settlements, geographical regions, countries, continents and words derived from them (relational adjectives and inhabitant names). Besides that, it also contains common abbreviations of institutions' names. This resource was created in a semi-automatic way, by first extracting the words and their forms from SJP.PL \footnote{\url{https://sjp.pl}} - a free, community driven dictionary of Polish - and then manually filtering out mismatched words. Our methodology for creating the lexicon was as follows: we searched the list of word definitions for occurrences of pre-defined terms i.g. for  inhabitants the search terms were "obywatel" ("citizen") and "mieszkaniec" ("inhabitant"), for relational adjectives we searched for "przymiotnik" ("adjecive"), "związany z" ("related to"), "pochodzący od" ("derived from") and selected only those definitions that also contained a name of settlement or country, for abbreviations we picked all short words written in capitals. The dictionary is relatively small compared to other resources we used - it contains about 10,000 words and more than 100,000 word forms.
\end{itemize}

\section{Wikipedia integration}
\label{sec:wikipedia}
One of the distinctive features of our system is the use the Wikipedia as a supplementary resource that helps to detect named entities, improving the accuracy of the model. Unlike lexicons and other static resources, modules based on evolving knowledge-bases such as Wikipedia can be automatically updated which is particularly important for named entity recognition systems that need to stay up to date in order to prevent the degradation of their performance over time. Our approach to utilizing Wikipedia involved labeling the articles with tags corresponding to named entity categories. We then used an entity linking method from the Wikipedia Miner \citep{milne2013open} toolkit, which we modified slightly to make it suitable for named entity recognition. The linked entities were assigned labels from our data set and the labels were used as an input to the deep learning model. In order to prepare the labeled set, we developed WikiMapper - an open source GUI application for tagging Wikipedia's pages \footnote{\url{https://github.com/sdadas/wiki-mapper}}. In this section, we describe the principles of operation of the aforementioned tool, the process of creating the data set and our method of improving named entity recognition by the use of an additional entity linking model.
\subsection{Data set preparation}
\begin{figure}
  \centering
  \includegraphics[scale=0.8]{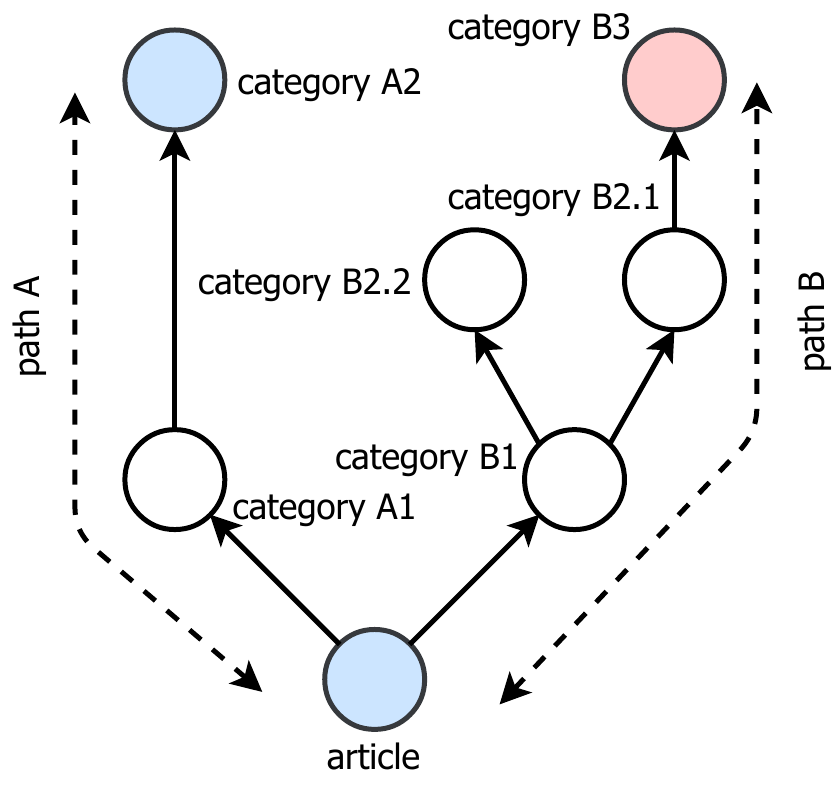}
  \caption{A graph showing the principle of label propagation in WikiMapper tool. The article inherits blue label from category A2 instead of red label from category B3 since the path A has length of 2 and the path B has length of 3.}
  \label{fig:wikimapper}
\end{figure}

The process of manually labeling data sets with millions of samples is usually costly and time-consuming. In some cases, however, it is possible to exploit the structure of data in order to make it more efficient. With regards to Wikipedia, we can take advantage of the fact that the structure is hierarchical i.e. each article is assigned at least one category and each category except the main categories has at least one parent category. When labeling the data, we rarely can benefit from moving down to the level of individual articles, we typically want to assign a label to all articles in a category. Depending on the task, this category can be more specific or general. On the basis of these assumptions, we have developed WikiMapper tool, with the objective of accelerating the workflow of manually tagging Wikipedia. It allows to quickly search articles and categories by title and efficiently traverse the category tree. We can use it to tag an individual article, a category or a set of pages matching specific search criteria. When a label is assigned to a category, it is automatically propagated to its child categories and articles. Since it's legal for an article or category to have multiple parents, every node in a graph can possibly have many inherited labels. In such cases, label conflicts are resolved in the following way: 1) For every path leading from unlabelled node to all of its labelled parent nodes, select the label connected with the shortest path. 2) When there are multiple paths of the same shortest length, select the most frequently occurring label among those paths. In the case of a tie, select the label arbitrarily. The label propagation method is illustrated in Figure \ref{fig:wikimapper}.

The labeled set has been prepared from the official dump of Polish Wikipedia \footnote{\url{https://dumps.wikimedia.org}}, containing over 1.2 million articles. During the labeling process, we exploited the regularities of category names in the Wikipedia. For instance, many of the urban objects are grouped in city related categories such as \emph{"Streets in [city\_name]"} or \emph{"Parks in [city\_name]"}. Most of the articles related to people, on the other hand, have a \emph{"Born in [year]"} or \emph{"Died in [year]"} category. Similar regularities can be found for other labels as well. As a result, it was possible to tag more than 70\% of articles in Wikipedia in less than an hour. Investing more time in preparing the data set leads to a better precision since we label smaller, more specific categories. For Polish Wikipedia, the results were satisfying after 5 hours of manual work, with 90\% of all articles covered. One of the advantages of our approach is the automatic labeling of new articles, due to the label propagation algorithm. Therefore, the labeled data set can be automatically expanded by downloading new Wikipedia dump. Moreover, WikiMapper is a general purpose software that may be used for preparing data sets for other supervised natural language processing tasks such as text classification, topic modelling, entity disambiguation among others. 

\subsection{Entity linking feature module}
Entity linking is the task of detecting and disambiguating named entities defined in a knowledge base (e.g. Wikipedia\footnote{\url{https://www.wikipedia.org}}, BabelNet\footnote{\url{https://babelnet.org}}, DBPedia\footnote{\url{https://wiki.dbpedia.org}}). Its objective is to create a mapping between mentions of named entities in text and their references in the knowledge base. This is closely related to the task of named entity recognition that can be seen as creating a mapping between instances of named entities and a set of pre-defined labels. Therefore it is possible to utilize entity linking model as a NER model by assigning a set of named entity labels to objects in the knowledge base. Such standalone methods have been proposed before \citep{pohl2013knowledge}, our approach employs the knowledge-based model as a part of a deep learning named entity recognition system. The process works as follows. First, the entity linking model finds references to entities in text. Next, a label assigned to such reference - coming from a data set described earlier - is used to tag this instance of named entity. Labels are then transformed to a one-hot vectors and passed as an input to the neural model.  

To explain the entity linking method itself, we need to summarize the methodology of the Wikipedia Miner \citep{milne2013open} framework. Two important definitions that we will use in this section is the \emph{concept} and the \emph{label}. The concept is an entity existing in the knowledge-base i.e. an article in Wikipedia. The label is a fragment of text that might be referring to a concept. Not all labels are unambiguous, some are used to describe different concepts depending on the context in which they appear. Wikipedia Miner collects several statistics from the dump of Wikipedia, one of them is a set of labels linking to each of the concepts with their frequencies. Additionally, the framework computes prior link probability $p(l)$ for every label $l$ which is the ratio of the number of times the label is used as a link between concepts to a total number of occurrences of this label in text corpus. For label and concept pairs $(l,c)$, a prior sense probability $ps(l,c)$ is defined as a ratio of the number of times this label has been used to refer to that specific concept, to the total number of times the label has been used as a link: 

\begin{equation}
    ps(l,c)=\frac{|\{(l,c')\: |\: c'=c, (l,c') \in L(l)\}|}{|L(l)|}
\end{equation}

Where $ps(l,c)$ is the prior sense probability of label-concept pair $(l,c)$ and $L(l)$ is the set of all links from the specific label $l$ to any concept $c'$ i.e. the set of all pairs $(l,c')$ present in the knowledge base.

The fundamental measure on which the entity linking method is based is the \emph{relatedness}. It reflects how closely two concepts are related and is based on Normalized Google Distance \citep{cilibrasi2007google}:

\begin{equation}
    rel(a,b)=1-\frac{log(max(|A|,|B|))-log(|A \cap B|)}{log(|W|)-log(min(|A|,|B|))}
\end{equation}

Where $rel(a,b)$ it the relatedness between concepts $a$ and $b$, $|A|$ is the number of articles linking to $a$, $|B|$ is the number of articles linking to $b$, $|A \cap B|$ is the number of articles linking to both $a$ and $b$, $|W|$ is the number of all articles in Wikipedia. 

Entity linking in Wikipedia Miner \citep{milne2013open} works in the context of a single text document. First, all candidate labels are found in the text. Labels with prior probability $p(l)$ lower than a specified threshold are discarded. From the remaining labels a set of possible concepts is determined. Next, a decision tree classifier called \emph{disambiguator} is used to score every label-concept pair $(l,c)$ found in the text. The model takes as an input three statistics related to the $(l,c)$ pair and outputs a single value from 0 to 1 representing the disambiguation probability i.e. the probability that the label $l$ is referring to the concept $c$. In order to introduce disambiguator inputs, we need to define two measures - quality of concept and quality of document (context). The quality of concept $q(c)$ is a measure of how similar the concept $c$ is to other concepts in the same document and how probable are the references to it in the document:

\begin{equation}
    q(c)=\frac{\sum_{k=1}^{|L_{c}|} ps(l_{k},c)}{|L_{c}|} \frac{\sum_{i=1}^{|C|}rel(c,c_{i})}{|C|}
\end{equation}

Where $q(c)$ is the quality of concept, $|L_{c}|$ is the number of labels in the document referring to this concept, $l_{k}$ is the k-th label referring to the concept $c$, $ps(l_{k},c)$ is the prior sense probability of the k-th label and concept $c$, $|C|$ is the number of all unambiguous concepts in a document, $rel(c,c_{i})$ is the relatedness measure between concept $c$ and $c_{i}$. Quality of document (context) $Q(C)$ is the sum of all unambiguous concept qualities $q(c)$ in the document:

\begin{equation}
    Q(C)=\sum_{i=1}^{|C|}q(c_{i})
\end{equation}

The complete list of features the disambiguation model uses for prediction includes:
\begin{itemize}
    \item Prior sense probability of the label-concept $ps(l,c)$.
    \item Relatedness of the concept $c$ to document context, which is defined as an average of relatedness of concept $c$ to all unambiguous concepts in the document weighted by the quality of concept:
    \begin{equation}
    relC(c,C)=\frac{\sum_{i=1}^{|C|}q(c_{i}) \: rel(c,c_{i})}{Q(C)}
    \end{equation}
     Where $relC(c,C)$ is the relatedness between concept $c$ and its document context $C$, $|C|$ is the number of all unambiguous concepts in a document, $q(c_{i})$ is the quality of concept $c_{i}$, $rel(c,c_{i})$ is the relatedness between concepts $c$ and $c_{i}$.
    \item Quality of document context $Q(C)$.
\end{itemize}

The original method filtered concepts based on disambiguator outputs, preserving only the highest scoring concepts. Next, an additional decision tree model named \emph{link detector} scored concepts for their relavance and importance. Finally, the most relevant concepts were selected to describe the document. In order to make the process more suited for named entity recognition, we decided to modify the concept selection method. We have observed that labels chosen by the algorithm are not always named entities themselves but phrases synonymous or related to named entities. Meanwhile, the guidelines for the named entity recognition are very strict on what should be considered as an entity. Since the scope of terms recognized by this method was too broad for our needs, we employed additional heuristics to discard label-concept pairs that were irrelevant to the task. This set of heuristic rules is applied as a pre-filtering step, before disambiguator scores are taken into account. We discard all the the pairs that meet one or more of the following conditions: 1) Label contains single character only. 2) Label is numeric (including Roman and Arabic numerals). 3) Label is a person name (based on a lexicon of people names) but the concept isn't tagged as \emph{persName}. We noticed a common ambiguity for such cases since many geographic objects, urban objects or organizations are named after a person. However, usually person names found in text refer directly to that person. Other cases are so rare that we decided to completely remove all non-person entity links matching this rule. 4) Label contains only lowercase common words or digits. We check whether a word is common using PoliMorf \citep{wolinski2012polimorf} dictionary.\\
Application of these rules could still leave us with labels having more than one concept. We could use disambiguator output to select concept with the highest score but as we have noticed, in practice taking an average of disambiguator output and relatedness to context $relC(c,C)$ yields better results in our case. Additionally, since our objective is to detect as many named entities as possible, we do not use link detector model for final selection of best concepts. All labels that passed the heuristic rules and are referring to a concept with assigned named entity class, are recognized as entities in the document.

\section{System evaluation}
\label{sec:evaluation}
In this section, we evaluate our named entity recognition system on the data from Poleval 2018 competition. PolEval is an evaluation series for natural language processing in Polish inspired by SemEval\footnote{\url{https://en.wikipedia.org/wiki/SemEval}}. PolEval 2018, which was held from June to August, included an evaluation of named entity recognition tools. A new annotated data set has been prepared specifically for this task in accordance to the guidelines of NKJP. In this competition, the original NKJP set has been used for training while the new data set for evaluation of the models.

\paragraph{Training and evaluation procedure}
Our model was trained on the annotated \emph{one million subcorpus} of NKJP. The publicly available version of the subcorpus is already tokenized and includes additional metadata such as lemmas or part-of-speech tags. On the other hand, the PolEval data set has been published in a raw text form. For this reason, it was required to tokenize and lemmatize the data before it could be used for model evaluation. For sentence splitting and tokenization, we used the tokenizer from LanguageTool \citep{milkowski2010developing}. For lemmatization, a simple frequency-based lemmatizer was used, selecting the most frequently occurring form in the corpus from the list of word forms suggested by  Morfologik\footnote{\url{https://github.com/morfologik/morfologik-stemming}}. Each of our models was trained for 10 epochs using NADAM optimization algorithm \citep{dozat2016incorporating} with the learning rate of 0.002, $\beta_1$ of 0.9, $\beta_2$ of 0.999 and batch size of 32. The models were evaluated on the PolEval corpus using the official evaluation script provided by the organizers.

\paragraph{Hyperparameter optimization} 
We used 10-fold cross-validation procedure on the NKJP data set in order to select optimal hyperparameters for our model. Various aspects of the sequence tagging architecture have been tested, including: 1) Type (LSTM\citep{hochreiter1997long}, GRU\citep{chung2014empirical}), size and number of recurrent layers. 2) Type and amount of dropout regularization. 3) The benefits of including Wikipedia and other external resources for Polish in the model. 4) The effect of using different word embeddings (Word2Vec\citep{mikolov2013distributed}, ELMo\citep{peters2018deep} or both models concatenated). 5) The effect of using character level word encoding techniques (CharCNN \citep{ma2016end, chiu2016named, peters2017semi}, CharBiLSTM \citep{lample2016neural, rei2017semi, liu2017empower} or none). 6) Type of the output layer (CRF, softmax). \\
Our final architecture used in the evaluation stage included three LSTM layers with 100 hidden units and a CRF inference layer. Variational dropout\citep{gal2016theoretically} with the probability of 0.25 has been applied before each recurrent layer. A character level representation based on CharCNN was selected for the final model.

\paragraph{Evaluated models}
In order to highlight the elements of our model that had the greatest impact on the performance in named entity recognition, we present four variations of the model. The best performing model utilizes contextualized ELMo embeddings by \citet{peters2018deep} trained on a large corpus of Polish and includes external features from Wikipedia and lexicons. As a baseline, we decided to evaluate the same architecture trained with static Word2Vec embeddings. Additionally, those two approaches were trained as standalone neural models, without using any external resources. 

\paragraph{Discussion}
We compare the performance of our system with three best submissions from PolEval 2018 in Table \ref{tab:eval_poleval}, reporting an absolute and relative improvement over the winning solution. The competing approaches are the following:
\begin{itemize}
\item \textbf{Per group LSTM-CRF with Contextual String Embeddings} \citep{borchmann2018lstmcrf} - The winning solution proposed an approach similar to \citet{akbik2018coling}, utilizing contextual embeddings based on shallow pre-trained language model in combination with GloVe word embeddings. To address the problem of overlapping named entities, they trained a few separate NER models, each dedicated for detecting only a subset of NKJP labels.
\item \textbf{PolDeepNer} \citep{marcinczuk2018poldeepner} - The results were produced using an ensemble of three different deep learning models.
\item \textbf{Liner2} \citep{marcinczuk2017liner2} - A CRF based named entity recognition system actively developed since 2013.
\end{itemize}
The evaluation results show that the contextual word embeddings and knowledge-based features both significantly increase the performance of the model. Of our four models, only the basic version with  Word2Vec and no external resources does not improve the score over the winning approach from PolEval. It's interesting to compare our standalone ELMo model with \citet{borchmann2018lstmcrf} since both utilize recently introduced contextual embeddings. Named entity recognition models with Flair\citep{akbik2018coling} and ELMo\citep{peters2018deep} have already been compared for English. Contrary to English results, ELMo seems to perform better than Flair for Polish. This can be due to the the fact that ELMo, with more layers and more parameters, is better suited for complex and challenging languages such as Polish. On the other hand, it may need to be pre-trained on a bigger corpora. We can also observe that knowledge-based features have a bigger impact on the model with static embeddings. For Word2Vec, there is a 2.7 percent absolute improvement in the score, while the improvement for ELMo is 1.7 percent. However, this is understandable, since the benefits of contextual information and external knowledge are overlapping. Compared with our baseline model, ELMo and language resources combined increase the score of our model by as much as 5 percent, from 84.6 to 89.6. Compared with the best solution from PolEval, our model improves the score by 3 percent, reducing the error rate by 22.4\%.

\begin{table}
  \centering
  \caption{Evaluation of our model on the data from PolEval 2018 Named Entity Recognition task. We compare for variations of our model with three best models from this competition. The final score for this task is based on the combination of \emph{exact match} and \emph{overlap} F1 scores (\emph{Final score = 0.8 * Overlap score + 0.2 * Exact match score}). For computing our scores, we used the official evaluation script provided by the organizers.}
  \aboverulesep=0ex
  \belowrulesep=0ex
  \begin{tabular}{l|clcc}
    \toprule
    Model & Final score & Improvement & Exact score & Overlap score \\
    \hline
    \makecell[l]{Liner2 \citep{marcinczuk2017liner2}} & 81.0 &  & 77.8 & 81.8 \\
    \makecell[l]{PolDeepNer \citep{marcinczuk2018poldeepner}} & 85.1 &  & 82.2 & 85.9 \\
    \makecell[l]{Per group LSTM-CRF with\\Contextual String Embeddings \citep{borchmann2018lstmcrf}} & 86.6 &  & 82.6 & 87.7 \\
    \hline
    \textbf{Without external resources} \\
    Our model (Word2Vec) & 84.6 & -2.0 / -12.9\% & 80.2 & 85.7 \\
    Our model (ELMo) & 87.9 & 1.3 / 9.7\% & 84.5 & 88.8 \\
    \hline
    \textbf{With Wikipedia and lexicons} \\
    Our model (Word2Vec) & 87.3 & 0.7 / 5.2\% & 82.9 & 88.4 \\
    Our model (ELMo) & \textbf{89.6} & 3.0 / 22.4\% & \textbf{86.2} & \textbf{90.5} \\
    \bottomrule
  \end{tabular}
  \label{tab:eval_poleval}
\end{table}

\section{Conclusions}
In this paper, we presented a neural architecture for named entity recognition and demonstrated how to improve its performance by using an entity linking model with a knowledge base such as Wikipedia. We have shown how to exploit the structure of Wikipedia to efficiently create labeled data sets for supervised natural language tasks. We evaluated selected elements of name entity recognition system including recurrent layers, external language resources, word based and character based representations. The evaluation of our system on data from PolEval 2018 shows that it can produce state-of-the-art performance for named entity recognition tasks in Polish. An improvement over previous approaches is the effect of combining knowledge-based features, contextual word embeddings and optimal hyperparameter selection. Integrating entity linking methods into NER systems is a promising direction that can be pursued to further improve the accuracy of such systems, especially using modern entity linking approaches involving deep learning methods such as graph embeddings that can be easily incorporated into other neural architectures.

\paragraph{Source code and language resources} Experimental results shown in this study have been produced using the library for Python programming language which is available at: \url{https://github.com/sdadas/yast}. Language resources and pre-trained models needed to replicate our results are available at \url{https://github.com/sdadas/polish-nlp-resources}.  

\bibliographystyle{unsrt}
\bibliography{references}

\begin{thebibliography}{40}
\providecommand{\natexlab}[1]{#1}
\providecommand{\url}[1]{\texttt{#1}}
\expandafter\ifx\csname urlstyle\endcsname\relax
  \providecommand{\doi}[1]{doi: #1}\else
  \providecommand{\doi}{doi: \begingroup \urlstyle{rm}\Url}\fi

\bibitem[Akbik et~al.(2018)Akbik, Blythe, and Vollgraf]{akbik2018coling}
Alan Akbik, Duncan Blythe, and Roland Vollgraf.
\newblock Contextual string embeddings for sequence labeling.
\newblock In \emph{{COLING} 2018, 27th International Conference on
  Computational Linguistics}, pages 1638--1649, 2018.

\bibitem[Bojanowski et~al.(2017)Bojanowski, Grave, Joulin, and
  Mikolov]{bojanowski2017enriching}
Piotr Bojanowski, Edouard Grave, Armand Joulin, and Tomas Mikolov.
\newblock Enriching word vectors with subword information.
\newblock \emph{Transactions of the Association for Computational Linguistics},
  5:\penalty0 135--146, 2017.

\bibitem[Borchmann et~al.(2018)Borchmann, Gretkowski, and
  Grali{\'n}ski]{borchmann2018lstmcrf}
{\L}ukasz Borchmann, Andrzej Gretkowski, and Filip Grali{\'n}ski.
\newblock Approaching nested named entity recognition with parallel lstm-crfs.
\newblock In \emph{Proceedings of AI and NLP Workshop Day 2018}, 2018.

\bibitem[Brahma(2018)]{brahma2018unsupervised}
Siddhartha Brahma.
\newblock Unsupervised learning of sentence representations using sequence
  consistency.
\newblock \emph{arXiv preprint arXiv:1808.04217}, 2018.

\bibitem[Camacho-Collados and Pilehvar(2018)]{camacho2018word}
Jose Camacho-Collados and Taher Pilehvar.
\newblock From word to sense embeddings: A survey on vector representations of
  meaning.
\newblock \emph{arXiv preprint arXiv:1805.04032}, 2018.

\bibitem[Chiu and Nichols(2016)]{chiu2016named}
Jason~PC Chiu and Eric Nichols.
\newblock Named entity recognition with bidirectional lstm-cnns.
\newblock \emph{Transactions of the Association for Computational Linguistics},
  4:\penalty0 357--370, 2016.

\bibitem[Chung et~al.(2014)Chung, Gulcehre, Cho, and
  Bengio]{chung2014empirical}
Junyoung Chung, Caglar Gulcehre, KyungHyun Cho, and Yoshua Bengio.
\newblock Empirical evaluation of gated recurrent neural networks on sequence
  modeling.
\newblock \emph{arXiv preprint arXiv:1412.3555}, 2014.

\bibitem[Cilibrasi and Vitanyi(2007)]{cilibrasi2007google}
Rudi~L Cilibrasi and Paul~MB Vitanyi.
\newblock The google similarity distance.
\newblock \emph{IEEE Transactions on knowledge and data engineering},
  19\penalty0 (3), 2007.

\bibitem[Collobert and Weston(2008)]{collobert2008unified}
Ronan Collobert and Jason Weston.
\newblock A unified architecture for natural language processing: Deep neural
  networks with multitask learning.
\newblock In \emph{Proceedings of the 25th international conference on Machine
  learning}, pages 160--167. ACM, 2008.

\bibitem[Collobert et~al.(2011)Collobert, Weston, Bottou, Karlen, Kavukcuoglu,
  and Kuksa]{collobert2011natural}
Ronan Collobert, Jason Weston, L{\'e}on Bottou, Michael Karlen, Koray
  Kavukcuoglu, and Pavel Kuksa.
\newblock Natural language processing (almost) from scratch.
\newblock \emph{Journal of Machine Learning Research}, 12\penalty0
  (Aug):\penalty0 2493--2537, 2011.

\bibitem[dos Santos et~al.(2015)dos Santos, Guimaraes, Niter{\'o}i, and
  de~Janeiro]{dos2015boosting}
C{\i}cero dos Santos, Victor Guimaraes, RJ~Niter{\'o}i, and Rio de~Janeiro.
\newblock Boosting named entity recognition with neural character embeddings.
\newblock In \emph{Proceedings of NEWS 2015 The Fifth Named Entities Workshop},
  page~25, 2015.

\bibitem[Dozat(2016)]{dozat2016incorporating}
Timothy Dozat.
\newblock Incorporating nesterov momentum into adam.
\newblock \emph{ICLR Workshop}, 2016.

\bibitem[Gal and Ghahramani(2016)]{gal2016theoretically}
Yarin Gal and Zoubin Ghahramani.
\newblock A theoretically grounded application of dropout in recurrent neural
  networks.
\newblock In \emph{Advances in neural information processing systems}, pages
  1019--1027, 2016.

\bibitem[Grali{\'n}ski et~al.(2009)Grali{\'n}ski, Jassem, Marci{\'n}czuk, and
  Wawrzyniak]{gralinski2009named}
Filip Grali{\'n}ski, Krzysztof Jassem, Micha{\l} Marci{\'n}czuk, and Pawe{\l}
  Wawrzyniak.
\newblock Named entity recognition in machine anonymization.
\newblock \emph{Recent Advances in Intelligent Information Systems}, 2009.

\bibitem[Hochreiter and Schmidhuber(1997)]{hochreiter1997long}
Sepp Hochreiter and J{\"u}rgen Schmidhuber.
\newblock Long short-term memory.
\newblock \emph{Neural computation}, 9\penalty0 (8):\penalty0 1735--1780, 1997.

\bibitem[Huang et~al.(2015)Huang, Xu, and Yu]{DBLP:journals/corr/HuangXY15}
Zhiheng Huang, Wei Xu, and Kai Yu.
\newblock Bidirectional {LSTM-CRF} models for sequence tagging.
\newblock \emph{CoRR}, abs/1508.01991, 2015.

\bibitem[Lample et~al.(2016)Lample, Ballesteros, Subramanian, Kawakami, and
  Dyer]{lample2016neural}
Guillaume Lample, Miguel Ballesteros, Sandeep Subramanian, Kazuya Kawakami, and
  Chris Dyer.
\newblock Neural architectures for named entity recognition.
\newblock In \emph{Proceedings of NAACL-HLT}, pages 260--270, 2016.

\bibitem[Liu et~al.(2017)Liu, Shang, Xu, Ren, Gui, Peng, and
  Han]{liu2017empower}
Liyuan Liu, Jingbo Shang, Frank Xu, Xiang Ren, Huan Gui, Jian Peng, and Jiawei
  Han.
\newblock Empower sequence labeling with task-aware neural language model.
\newblock \emph{arXiv preprint arXiv:1709.04109}, 2017.

\bibitem[Ma and Hovy(2016)]{ma2016end}
Xuezhe Ma and Eduard Hovy.
\newblock End-to-end sequence labeling via bi-directional lstm-cnns-crf.
\newblock In \emph{Proceedings of the 54th Annual Meeting of the Association
  for Computational Linguistics (Volume 1: Long Papers)}, volume~1, pages
  1064--1074, 2016.

\bibitem[Marci{\'n}czuk and Janicki(2012)]{marcinczuk2012optimizing}
Micha{\l} Marci{\'n}czuk and Maciej Janicki.
\newblock Optimizing {CRF}-based model for proper name recognition in polish
  texts.
\newblock In \emph{International Conference on Intelligent Text Processing and
  Computational Linguistics}, pages 258--269. Springer, 2012.

\bibitem[Marci{\'n}czuk et~al.(2013)Marci{\'n}czuk, Koco{\'n}, and
  Janicki]{marcinczuk2013liner2}
Micha{\l} Marci{\'n}czuk, Jan Koco{\'n}, and Maciej Janicki.
\newblock Liner2--a customizable framework for proper names recognition for
  polish.
\newblock In \emph{Intelligent tools for building a scientific information
  platform}, pages 231--253. Springer, 2013.

\bibitem[Marci{\'n}czuk et~al.(2017)Marci{\'n}czuk, Koco{\'n}, and
  Oleksy]{marcinczuk2017liner2}
Micha{\l} Marci{\'n}czuk, Jan Koco{\'n}, and Marcin Oleksy.
\newblock Liner2 - a generic framework for named entity recognition.
\newblock In \emph{Proceedings of the 6th Workshop on Balto-Slavic Natural
  Language Processing}, pages 86--91, 2017.

\bibitem[Marci{\'n}czuk et~al.(2018)Marci{\'n}czuk, Koco{\'n}, and
  Gawor]{marcinczuk2018poldeepner}
Michał Marci{\'n}czuk, Jan Koco{\'n}, and Michał Gawor.
\newblock Recognition of named entities for polish-comparison of deep learning
  and conditional random fields approaches.
\newblock In Maciej Ogrodniczuk and {\L}ukasz Kobyli{\'n}ski, editors,
  \emph{Proceedings of the PolEval 2018 Workshop}, pages 77--92. Institute of
  Computer Science, Polish Academy of Science, 2018.

\bibitem[Mikolov et~al.(2013)Mikolov, Sutskever, Chen, Corrado, and
  Dean]{mikolov2013distributed}
Tomas Mikolov, Ilya Sutskever, Kai Chen, Greg~S Corrado, and Jeff Dean.
\newblock Distributed representations of words and phrases and their
  compositionality.
\newblock In \emph{Advances in neural information processing systems}, pages
  3111--3119, 2013.

\bibitem[Mi{\l}kowski(2010)]{milkowski2010developing}
Marcin Mi{\l}kowski.
\newblock Developing an open-source, rule-based proofreading tool.
\newblock \emph{Software: Practice and Experience}, 40\penalty0 (7):\penalty0
  543--566, 2010.

\bibitem[Milne and Witten(2013)]{milne2013open}
David Milne and Ian~H Witten.
\newblock An open-source toolkit for mining wikipedia.
\newblock \emph{Artificial Intelligence}, 194:\penalty0 222--239, 2013.

\bibitem[Pennington et~al.(2014)Pennington, Socher, and
  Manning]{pennington2014glove}
Jeffrey Pennington, Richard Socher, and Christopher Manning.
\newblock Glove: Global vectors for word representation.
\newblock In \emph{Proceedings of the 2014 conference on empirical methods in
  natural language processing (EMNLP)}, pages 1532--1543, 2014.

\bibitem[Peters et~al.(2017)Peters, Ammar, Bhagavatula, and
  Power]{peters2017semi}
Matthew Peters, Waleed Ammar, Chandra Bhagavatula, and Russell Power.
\newblock Semi-supervised sequence tagging with bidirectional language models.
\newblock In \emph{Proceedings of the 55th Annual Meeting of the Association
  for Computational Linguistics (Volume 1: Long Papers)}, volume~1, pages
  1756--1765, 2017.

\bibitem[Peters et~al.(2018)Peters, Neumann, Iyyer, Gardner, Clark, Lee, and
  Zettlemoyer]{peters2018deep}
Matthew Peters, Mark Neumann, Mohit Iyyer, Matt Gardner, Christopher Clark,
  Kenton Lee, and Luke Zettlemoyer.
\newblock Deep contextualized word representations.
\newblock In \emph{Proceedings of the 2018 Conference of the North American
  Chapter of the Association for Computational Linguistics: Human Language
  Technologies, Volume 1 (Long Papers)}, volume~1, pages 2227--2237, 2018.

\bibitem[Piskorski(2005)]{piskorski2005named}
Jakub Piskorski.
\newblock Named-entity recognition for polish with sprout.
\newblock In \emph{Intelligent Media Technology for Communicative
  Intelligence}, pages 122--133. Springer, 2005.

\bibitem[Piskorski et~al.(2004)Piskorski, Sch{\"a}fer, and
  Xu]{piskorski2004shallow}
Jakub Piskorski, Ulrich Sch{\"a}fer, and Feiyu Xu.
\newblock Shallow processing with unification and typed feature
  structures--foundations and applications.
\newblock \emph{Knstliche Intelligenz}, 1\penalty0 (1), 2004.

\bibitem[Pohl(2013)]{pohl2013knowledge}
Aleksander Pohl.
\newblock Knowledge-based named entity recognition in polish.
\newblock In \emph{Computer Science and Information Systems (FedCSIS), 2013
  Federated Conference on}, pages 145--151. IEEE, 2013.

\bibitem[Przepi{\'o}rkowski et~al.(2012)Przepi{\'o}rkowski, Banko, G{\'o}rski,
  and Lewandowska-Tomaszczyk]{przepiorkowski2012narodowy}
Adam Przepi{\'o}rkowski, Miros{\l}aw Banko, Rafa{\l}~L G{\'o}rski, and Barbara
  Lewandowska-Tomaszczyk.
\newblock Narodowy korpus jezyka polskiego [eng.: National corpus of polish].
\newblock \emph{Wydawnictwo Naukowe PWN, Warsaw}, 2012.

\bibitem[Radziszewski(2013)]{radziszewski2013tiered}
Adam Radziszewski.
\newblock A tiered {CRF} tagger for polish.
\newblock In \emph{Intelligent tools for building a scientific information
  platform}, pages 215--230. Springer, 2013.

\bibitem[Rei(2017)]{rei2017semi}
Marek Rei.
\newblock Semi-supervised multitask learning for sequence labeling.
\newblock \emph{arXiv preprint arXiv:1704.07156}, 2017.

\bibitem[Reimers and Gurevych(2017)]{reimers2017optimal}
Nils Reimers and Iryna Gurevych.
\newblock Optimal hyperparameters for deep lstm-networks for sequence labeling
  tasks.
\newblock \emph{arXiv preprint arXiv:1707.06799}, 2017.

\bibitem[Tjong Kim~Sang and De~Meulder(2003)]{tjong2003introduction}
Erik~F Tjong Kim~Sang and Fien De~Meulder.
\newblock Introduction to the conll-2003 shared task: Language-independent
  named entity recognition.
\newblock In \emph{Proceedings of the seventh conference on Natural language
  learning at HLT-NAACL 2003-Volume 4}, pages 142--147. Association for
  Computational Linguistics, 2003.

\bibitem[Waszczuk(2012)]{waszuk2012nerf}
Jakub Waszczuk.
\newblock {NERF} - {N}amed entity recognition tool based on linear-chain
  {CRF}s.
\newblock \url{http://zil.ipipan.waw.pl/Nerf}, 2012.

\bibitem[Wolinski et~al.(2012)Wolinski, Milkowski, Ogrodniczuk, and
  Przepi{\'o}rkowski]{wolinski2012polimorf}
Marcin Wolinski, Marcin Milkowski, Maciej Ogrodniczuk, and Adam
  Przepi{\'o}rkowski.
\newblock Polimorf: a (not so) new open morphological dictionary for polish.
\newblock In \emph{LREC}, pages 860--864, 2012.

\bibitem[Yang et~al.(2018)Yang, Liang, and Zhang]{yang2018design}
Jie Yang, Shuailong Liang, and Yue Zhang.
\newblock Design challenges and misconceptions in neural sequence labeling.
\newblock \emph{arXiv preprint arXiv:1806.04470}, 2018.

\end{thebibliography}

\clearpage
\appendix
\section{Appendix: Hyperparameter tuning}

\label{sec:hyperparameter_tuning}
In this section, we describe aspects of the process of hyperparameter tuning for our model. For the initial configuration, we followed the advice of recent publications on the topic of hyperparameter selection for sequence labeling tasks \citep{reimers2017optimal,yang2018design}. Then, we conducted a series of experiments in order to fine-tune the model for the problem of named entity recognition in Polish. In most cases, initially chosen hyperparameters turned out to be optimal or near-optimal. We have decided to introduce only minor changes to the configuration that we explain in this section. Additionally, we evaluate the usefulness of natural language resources used by our model i.e. lexicons and the Wikipedia module. Finally, we evaluate models for text representation, including recently introduced deep contextualized word representations \citep{peters2018deep}.

Our deep learning model has a number of inputs suited for specific input modules. Text representation modules usually take whole word, its base form or a vector of characters forming the word as an input. One-hot modules are external to the model - they process the text and pass pre-computed vector features that can be directly used by the recurrent layers. Initially, we included in the model all one-hot features described earlier and two text based modules - a pre-trained Word2Vec embedding model for Polish and a character based module based on bidirectional LSTM. The weights of Word2Vec model were fixed during training. The weights of character based module were randomly initialized and learned from scratch. Since the word embedding model has been pre-trained on lemmatized corpus, we passed lower case base form of a word as its input. Character based module, on the other hand, took original word with preserved casing. Later, we included ELMo \citep{peters2018deep} model which used original word forms as well. As mentioned in Section \ref{sec:architecture}, the complete word representation is a concatenation of text based vectors and one-hot vectors. This representation passes through a number of bidirectional recurrent layers. Our initial configuration included 2 bidirectional LSTM layers with 100 recurrent units each. Dropout at the input of each LSTM layer has been set to 0.25. The output of the last recurrent layer is passed to the CRF based inference layer.

All experiments shown in this section were evaluated on the NKJP data set, the PolEval 2018 data set has been used only for the final evaluation of the model described in Section \ref{sec:evaluation}. If not stated otherwise, the default procedure for carrying out experiments was the following. We used 10-fold cross-validation to train the model for every set of examined hyperparameters, then we averaged the F1 scores computed for each of the folds. The reported F1 scores are computed in accordance with ConLL 2003 evaluation procedure \citep{tjong2003introduction}, counting only the exact matches of named entities. 

\subsection{Recurrent layers}
To determine the architecture of recurrent layers, we evaluated 16 different configurations. Two types of layers were tested (LSTM\citep{hochreiter1997long}, GRU\citep{chung2014empirical}), two cell sizes (100 and 200 units) and architectures with one to four stacked recurrent layers. Our results are shown in Table \ref{tab:eval_layers}. As noted by \citet{reimers2017optimal}, for English CoNLL 2003 named entity recognition task, an optimal configuration includes two LSTM layers and cell size of around 100 units. That was our initial configuration for Polish. However, the experiment showed that for LSTM the model performed better on average with three layers. We assume there are two reasons for this discrepancy. Firstly, the definitions of named entities from NKJP are more ambiguous and more context dependent that those from CoNLL. It is common for the same sequence of words to be assigned different classes based on the sentence or document context e.g. whether it is a modern or historical text. Secondly, the morphological structure of Polish language is more complex than English. We also noticed a little improvement in F1 scores after increasing the number of recurrent untis to 200, although the difference was not significant. This observation is consistent with the results of \citet{reimers2017optimal} who found that the number of layers had a greater impact on performance than the size of single layer. Eventually, we decided to set the number of layers to three but leave the LSTM size at 100 units. Next, for the selected recurrent architecture we evaluated four levels of variational dropout at the input of each recurrent layer from 0 to 0.5. The results of this experiment are shown in Table \ref{tab:dropout}. Our initial configuration with dropout of 0.25 turned out to be the best of the set of tested values. We also compared two popular tagging schemes used in sequence tagging problems - BIO and BIOES, but our experiments have shown no significant performance difference. Therefore, for all the experiments in this section we used the BIO tagging scheme.

\begin{table}
  \centering
  \caption{Comparison of the type of recurrent layer, number of recurrent layers and cell size (number of recurrent units). Each number is an average F1 score of 10-fold cross validation on NKJP dataset. An absolute difference to the best score is shown in parentheses.}
  \aboverulesep=0ex
  \belowrulesep=0ex
  \setlength\extrarowheight{7pt}
  \begin{tabular}{l|c|cccc}
    \toprule
    \multirow{2}{*}{Layer type} & \multirow{2}{*}{Cell size} & \multicolumn{4}{c}{\makecell{Number of layers}} \\
    \cmidrule(r){3-6}
    & & 1 & 2 & 3 & 4 \\
    \hline
    \multirow{2}{*}{LSTM} & 100 & \makecell{86.98\\(-0.98)} & \makecell{87.83\\(-0.13)} & \makecell{87.94\\(-0.03)} & \makecell{87.61\\(-0.36)} \\
     & 200 & \makecell{87.12\\(-0.85)} & \makecell{87.85\\(-0.12)} & \makecell{\textbf{87.97}} & \makecell{87.73\\(-0.24)} \\
     \hline
     \multirow{2}{*}{GRU} & 100 & \makecell{86.11\\(-1.86)} & \makecell{85.83\\(-2.13)} & \makecell{85.49\\(-2.47)} & \makecell{84.27\\(-3.70)}\\
     & 200 & \makecell{85.59\\(-2.38)} & \makecell{84.92\\(-3.04)} & \makecell{83.69\\(-4.26)} & \makecell{82.62\\(-5.35)} \\
    \bottomrule
  \end{tabular}
  \label{tab:eval_layers}
\end{table}

\begin{table}
  \centering
  \caption{Comparison of dropout percentage for the input of LSTM layers. Each number is an average F1 score of 10-fold cross validation on NKJP dataset. An absolute difference to the best score is shown in parentheses.}
  \aboverulesep=0ex
  \belowrulesep=0ex
  \begin{tabular}{l|cl}
    \toprule
    Dropout & F1-score \\
    \hline
	0.00 & 86.72 & (-1.23) \\
    0.10 & 87.57 & (-0.36) \\
    0.25 & \textbf{87.93} & \\
    0.50 & 85.75 & (-2.18)\\
    \bottomrule
  \end{tabular}
  \label{tab:dropout}
\end{table}

\subsection{One-hot features}
Our named entity recognition model relies on a number additional one-hot encoded features which have been summarized in Section \ref{sec:architecture}. Apart from the module based on Wikipedia, we use five lexicons. Two of them have been created by us \footnote{The lexicons are publicly available and can be downloaded from \url{https://github.com/sdadas/polish-nlp-resources}}. We prapared the \emph{Extras} lexicon specifically for this task while the \emph{PersonNames} has been used by us before for other applications related to natural language processing. Three other dictionaries have been publicly available, some for more that 10 years and have been utilized by various named entity recognition tools. In this subsection we explore the impact of the lexicons on the performance of named entity recognition system. In the first experiment, we trained our model restricting the use of one-hot features. We checked how the model performed without additional features, with the capitalization feature only and with capitalization feature combined with each of the external resources. Since our initial experiments suggested that the model cannot fully exploit these resources without the casing information, we included the capitalization feature for all instances. The results of the experiment are show in Table \ref{tab:feature_fscores}. We can see that largest improvement in F1 score with Wikipedia, 0.92 more than the model without one-hot features. PoliMorf and Extras dictionaries have a comparable impact. Other resources have a noticeably smaller effect than Wikipedia. It should be noted that the scope of each of the resources may overlap and the aggregated effect of combined features is smaller than the sum of individual performance improvements. For example, the inclusion of Wikipedia module improves the F1 score by 0.72 percent over the capitalization feature while the inclusion of all external resources increases the score by about 1.5 percent. To further investigate the usefulness of language resources we carried out another test in which we trained simple logistic regression models based on a single one-hot feature only. This experiment shows how each feature performs without any word or character based information. Table \ref{tab:feature_validation} contains precision, recall and F1 scores of logistic regression models by named entity category. The results reveal more details on how the different resources complement each other. Once again, Wikipedia seems to be strong in every category except temporal based entities. However, PoliMorf and Extras dictionaries outperform Wikipedia in \emph{persName} and \emph{placeName} categories respectively. The GPNE lexicon demonstrates reasonably good performance on those two classes as well. This is consistent with our first experiments where Wikipedia, Extras, PoliMorf and GPNE had a strongest impact on the F1 score. The reason for Wikipedia's efficiency is high recall in four named entity categories, especially in \emph{geogName} and \emph{orgName} where it is significantly higher than any other resource we used. Other lexicons usually specialize in one or two categories which requires to use a combination of several resources to cover the domain of the task.

\begin{table}
  \centering
  \caption{F1 scores of our model trained on a subset of one-hot features. For \emph{None}, no feature has been used, for \emph{Capitalization} we used only the capitalization feature, for all other cases a single lexicon feature has been used along with the capitalization feature as we noticed that the word casing information is essential to fully utilize most of the lexicon based features. Each number is an average F1 score of 10-fold cross validation on NKJP dataset. An absolute difference to the best score is shown in parentheses.}
  \aboverulesep=0ex
  \belowrulesep=0ex
  \begin{tabular}{l|cl}
    \toprule
    Features & F1-score &  \\
    \hline
	None & 86.23 & (-0.92) \\
    Capitalization & 86.43 & (-0.72) \\
    PersonNames & 86.59 & (-0.56) \\
    NELexicon2 & 86.67 & (-0.48) \\
    GPNE & 86.79 & (-0.36) \\
    PoliMorf & 86.96 & (-0.19) \\
    Extras & 87.01 & (-0.14) \\
    Wikipedia & \textbf{87.15} & \\
    \bottomrule
  \end{tabular}
  \label{tab:feature_fscores}
\end{table}

\begin{table}
  \centering
  \caption{Scores for logistic regression models trained on one-hot vectors generated by a single feature module only. The reported scores show an effect of the feature on each main named entity class. The first number in each cell is an F1-score of the logistic regression model, numbers in parentheses are precision and recall of the model.}
  \aboverulesep=0ex
  \belowrulesep=0ex
  \setlength\extrarowheight{7pt}
  \begin{tabular}{l|cccccc}
    \toprule
    Feature & date & geogName & orgName & persName & placeName & \makecell[{{p{1.5cm}}}]{\centering time} \\
    \hline
	PersonNames & 0 & 0 & 0 & \makecell{62.65\\(83.77 / 50.3)} & 0 & 0\\
    PoliMorf & 0 & 0 & \makecell{18.88\\(67.70 / 11.00)} & \textbf{\makecell{67.24\\(74.59 / 61.21)}} & \makecell{61.93\\(64.78 / 59.32)} & 0 \\
    Wikipedia & 0 & \textbf{\makecell{41.02\\(49.27 / 35.14)}} & \textbf{\makecell{47.81\\(67.97 / 36.87)}} & \makecell{64.10\\(85.22 / 51.37)} & \makecell{57.15\\(57.27 / 57.02)} & 0 \\
    GPNE & \textbf{\makecell{57.15\\(98.64 / 40.23)}} & \makecell{13.87\\(58.30 / 7.87)} & \makecell{5.88\\(66.29 / 3.08)} & \makecell{61.35\\(70.75 / 54.16)} & \makecell{56.03\\(83.34 / 42.20)} & 0 \\
    NELexicon2 & 0 & \makecell{4.24\\(69.97 / 2.19)} & \makecell{30.43\\(74.06 / 19.15)} & \makecell{49.28\\(57.18 / 43.30)} & \makecell{38.88\\(48.94 / 32.25)} & 0 \\
    Extras & 0 & \makecell{20.49\\(73.15 / 11.91)} & \makecell{17.78\\(82.80 / 9.96)} & 0 & \textbf{\makecell{62.01\\(78.32 / 51.32)}} & 0 \\
    \bottomrule
  \end{tabular}
  \label{tab:feature_validation}
\end{table}

\subsection{Text representations}
Since it has been shown \citep{collobert2011natural} that word embeddings play a fundamental role in improving performance in sequence tagging problems, they are commonly included in named entity recognition models. The popular choices for embedding layers are Word2Vec \citep{mikolov2013distributed}, GloVe \citep{pennington2014glove} and FastText \citep{bojanowski2017enriching}. Recently, a number of contextual word embedding methods based on language model pre-training have been proposed \citep{peters2017semi, peters2018deep, akbik2018coling} which proved to be beneficial for many natural language processing tasks including named entity recognition. In this work we investigate the performance of our model with Word2Vec embeddings, ELMo (Embedding from Language Models) and a combination of both. We choose to use ELMo over other contextual approaches since this method was thoroughly evaluated \citep{camacho2018word, brahma2018unsupervised, peters2018deep} despite being proposed only recently. In addition to word representations, some NER models utilize character based text representations. Two popular approaches are character level encoders based on CNNs \citep{ma2016end, chiu2016named, peters2017semi} and bidirectional LSTMs \citep{lample2016neural, rei2017semi, liu2017empower}. 

Table \ref{tab:eval_embeddings} illustrates the performance of our model using different word and character level embedding methods. A combination of three word representations (Word2Vec, ELMo, ELMo + Word2Vec) and three approaches to character based representation (CNN, LSTM and no representation) were evaluated. The word embedding models have been trained on corpus consisting of Polish books, articles and text extracted from Polish Wikipedia, 1.5 billion tokens at total. We trained Word2Vec model with vector size of 100 using negative sampling. The embedding includes words occurring at least three times in the training corpus and additionally all numbers from 0 to 10'000, a set of predefined punctuation symbols, forenames and surnames. The corpus was lemmatized and lowercased before training. For ELMo, we followed the training procedure of \citet{peters2018deep} and used the same configuration as their original model. In this case, unaltered corpus was used for training. For character level encoders, we used character embedding layer with 50 units. Character CNN included single convolutional layer with 3x3 kernel and 30 filters. Character LSTM included bidirectional LSTM layer with 100 hidden units. The evaluation results show a significant performance improvement when using ELMo. What is more, it seems that model with ELMo does not benefit from including additional word or character based representations. While there is a small performance gain from character embedding if only Word2Vec is used, this effect is no longer present in ELMo based models. This is expected as character information is already included in representations generated by ELMo. \citet{akbik2018coling} reported an improvement of F1 score on NER task by using concatenation of static and contextual word embeddings (GloVe + language model) but in our case combining Word2Vec with ELMo did not bring any significant performance improvement.

\begin{table}
  \centering
  \caption{Comparison of features based on text representations. We have tested three configurations for word representation (Word2Vec, ELMo and a combination of both models) and three configurations for character based representation (none, CharCNN and CharBiLSTM). Each number is an average F1 score of 10-fold cross validation on NKJP dataset. An absolute difference to the best score is shown in parentheses.}
  \setlength\extrarowheight{3pt}
  \aboverulesep=0ex
  \belowrulesep=0ex
  \setlength\extrarowheight{7pt}
  \begin{tabular}{l|ccc}
    \toprule
    \multirow{2}{*}{Word representation} & \multicolumn{3}{c}{Character representation} \\
    \cmidrule(r){2-4}
    & \makecell[{{p{1.7cm}}}]{\centering None} & \makecell[{{p{1.7cm}}}]{\centering CharCNN} & \makecell[{{p{1.7cm}}}]{\centering CharBiLSTM} \\
    \hline
    Word2Vec & \makecell{87.48\\(-2.26)} & \makecell{87.57\\(-2.17)} & \makecell{87.59\\(-2.15)} \\
    ELMo & \makecell{89.65\\(-0.09)} & \makecell{\textbf{89.74}} & \makecell{89.68\\(-0.06)} \\
    ELMo + Word2Vec & \makecell{89.65\\(-0.09)} & \makecell{89.70\\(-0.04)} & \makecell{89.68\\(-0.06)} \\
    \bottomrule
  \end{tabular}
  \label{tab:eval_embeddings}
\end{table}

\end{document}